# An Ordering Optimization for Fruit-Vegetable in High-Density Environment based on Genetic Algorithm

Jaehwi Seol [1,2], Jeonghyeon Pak[1,3], and Hyoung Il Son[1,3,4*]

**Abstract**

This paper presents a robotic harvesting method that combines RGB-D fruit localization, GA-based harvest ordering, and waypoint-based robot motion in dense and unstructured crop environments. The global RGB-D camera provides 3D fruit positions for ordering, while the end-effector-mounted camera provides close-range observation during execution. The GA determines the harvesting sequence, and static or dynamic waypoints define the intermediate robot motion. The method was evaluated in a controlled laboratory testbed and in four-, five-, and six-fruit field conditions. GA solutions were compared with exhaustive enumeration, and both waypoint offsets were examined by parameter sensitivity analysis. Continuous execution reduced unnecessary Home-return motion, while the dynamic waypoint changed with the remaining fruit distribution without changing the GA sequence. In the field evaluation, the dynamic-waypoint continuous strategy produced 40 collision events and a 70.7% target-approach success rate, compared with 55 collision events and 65.3% for the static-waypoint continuous strategy.

Jaehwi Seol: seol0810@jnu.ac.kr, ORCID 0000-0003-1382-4415,
Jeonghyeon Pak: jhpak@jnu.ac.kr, ORCID 0000-0003-1070-9681,
Hyoung Il Son: hison@jnu.ac.kr, ORCID 0000-0002-7249-907X.

[1] Department of Convergence Biosystems Engineering, Chonnam National University, 77 Yongbong-ro, Buk-gu, Gwangju 61186, Republic of Korea

[2] Kimchi Factory Research Group, World Institute of Kimchi, Gwangju, 61755, Republic of Korea

[3] Interdisciplinary Program in IT-Bio Convergence System, Chonnam National University, 77 Yongbong-ro, Buk-gu, Gwangju 61186, Republic of Korea

[4] Research Center for Biological Cybernetics, Chonnam National University, 77 Yongbong-ro, Buk-gu, Gwangju 61186, Republic of Korea

[*]Corresponding author

## 1. Introduction

Agricultural robots, which autonomously execute tasks and make real-time decisions in complex, unstructured environments, have emerged as a viable solution to labor shortages (Seol et al., 2024b). Recent advancements in agricultural automation have facilitated the development of a broad range of robotic platforms capable of core farming operations, such as harvesting (Jun et al., 2021; Park et al., 2023), pruning (Li et al., 2024), thinning (Du et al., 2025), and precision spraying (Seol et al., 2022a,b). These systems have increasingly been deployed to automate repetitive and labor-intensive tasks traditionally performed by humans. Beyond these operations, robotic applications extend to autonomous navigation in greenhouses (Pak et al., 2022), UAV-based insect localization and tracking (Pak et al., 2023, 2024), environmental mapping (Seol et al., 2024a), and irrigation monitoring (Pak and Son, 2025). These applications illustrate the use of robotic perception and autonomous operation across diverse agricultural settings. Recent agricultural automation has also integrated harvesting, unloading, and transportation into coordinated autonomous operation systems (Ding et al., 2025). Together, these developments support efforts to improve the productivity, safety, and longterm sustainability of agricultural production.

Harvesting tasks are challenged by the unstructured and cluttered nature of agricultural environments (Fig. 1), which are characterized by irregular plant arrangements, unstructured terrain, dense leaves, and non-uniform fruit distribution (Bhattarai et al., 2024). Manual harvesting in these settings requires considerable time and labor, often making the process inefficient and economically burdensome. The ongoing decline in the availability of agricultural labor intensifies these challenges, causing delays and potential post-ripening degradation, increasing the risk of economic loss. Accordingly, time-efficient, labor-saving robotic harvesting solutions that can operate reliably in field conditions are necessary.

Robotic harvesting concerns two primary manipulation tasks: grasping and cutting. Prior studies have focused on improving perception systems to support these operations, particularly in dense crop environments where precise fruit localization is critical. Advances in computer vision and learning-based detection have advanced object recognition and 3D localization, contributing to more reliable trajectory planning. However, field deployment remains limited due to persistent problems (e.g., occlusion, overlapping targets, and ambiguous depth cues). These factors can make the spatial relationship among targets difficult to determine from image-plane information alone. Figure 2 shows that targets with similar two-dimensional image positions can be located at different depths. Therefore, accurate 3D fruit positions are important for determining the harvesting order in dense scenes.

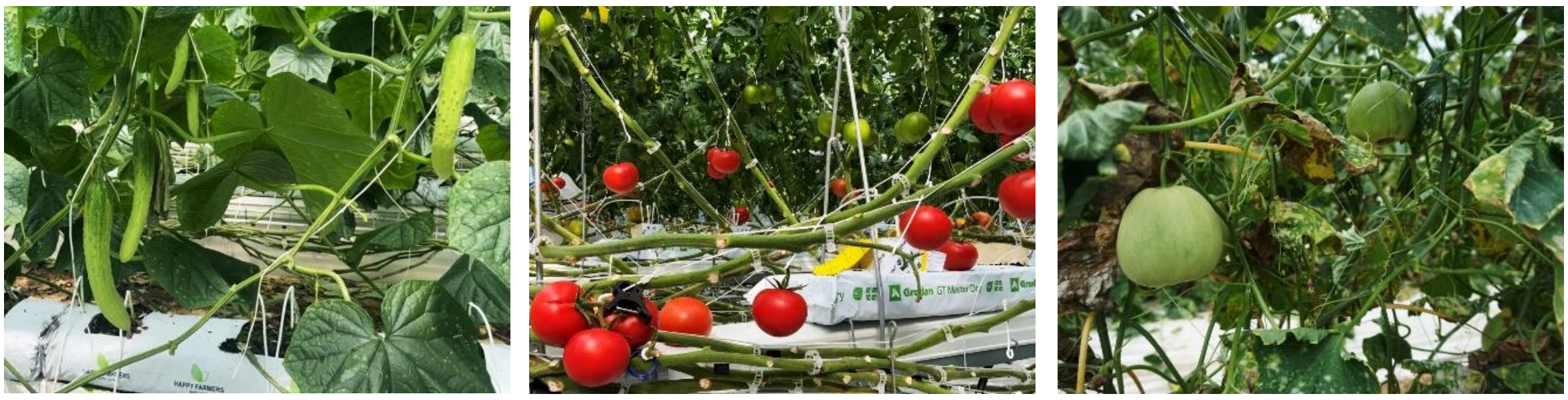

(a) (b) (c)

**Fig. 1** High-density fruit environments: (a) cucumber, (b) tomato, and (c) Korean melon.

A common approach to determine the harvesting order is to frame the task as a traveling salesman problem (TSP) based on Euclidean distances between fruit positions. Harvest ordering based on two-dimensional image coordinates can misrepresent inter-fruit distances when targets lie at different depths. RGB-D sensing provides the 3D position of each detected fruit, including its depth. In this study, the GA determines the harvesting sequence using the three-dimensional Euclidean distance among the detected fruit. Static and dynamic waypoints are then used to define the intermediate robot motion for executing the sequence. The aim is to improve motion efficiency and execution consistency in unstructured environments.

This paper proposes an integrated harvest-ordering framework that aims to improve efficiency and operational robustness in fruit harvesting under these conditions. The framework combines RGB-D fruit localization, GA-based harvest ordering, and waypoint-based robot motion. The GA determines the target order, while static and dynamic waypoints define the intermediate robot motion. The dynamic waypoint is updated from the remaining fruit without changing the GA sequence. The primary contributions of this paper are summarized below:

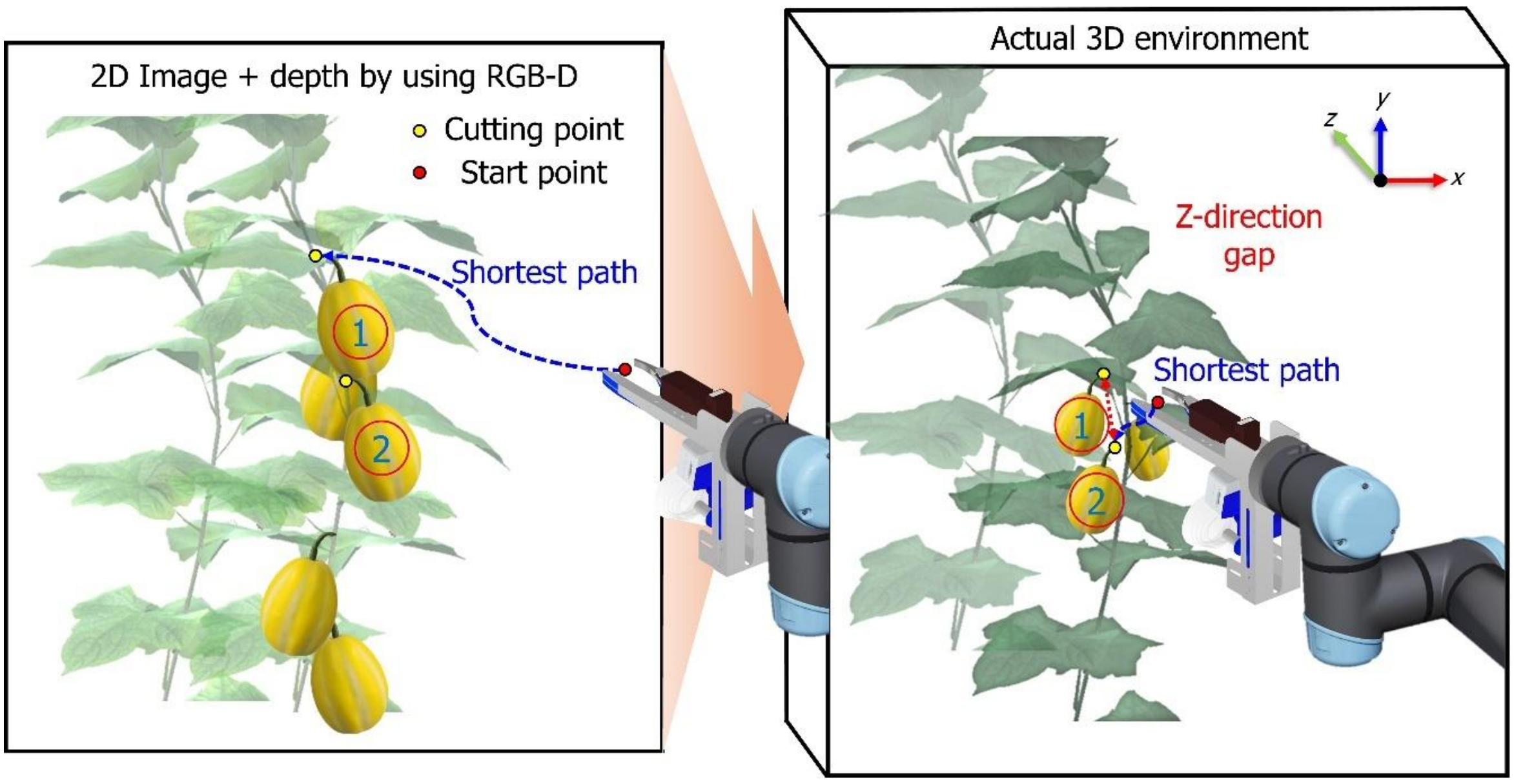

**Fig. 2** Problem definition: 2D path planning under depth uncertainty leads to deviation from the true 3D path in dense environments; left: 2D-planned path from RGB-D image; right: actual shortest path computed in 3D environment.

- A global–local RGB-D sensing configuration is developed to integrate robot base fruit localization for harvest-order optimization with close-range target observation during robot execution.

- A GA-based harvest-ordering method is proposed to integrate sequence optimization with static and dynamic waypoint strategies, adapting intermediate motion to the remaining-target distribution.

- The proposed framework is validated through comparative field experiments in unstructured agricultural environments, demonstrating the practical benefits of integrating harvest-order optimization with robot execution.

The remainder of this paper is organized as follows. Section 2 provides a comprehensive review of the related works. Section 3 elaborates on the fruit detection and localization perception system. Section 4 details the GA-based harvest ordering system. Section 5 outlines the experimental setup and presents the comparative evaluations of harvesting order algorithms implemented using a vision-based robotic manipulator under field conditions. Section 6 details the study limitations, and Section 7 concludes.

## 2. Related Works

### 2.1 Fruit perception

Detection performance in complex scenes (e.g., those with variable lighting and occlusion) has significantly improved with deep learning (DL) methods compared to traditional feature-based techniques. The DL-based object detectors typically comprise three components: a backbone for feature extraction, a neck for feature enhancement, and a head for classification and localization. These models are typically categorized as one- or two-stage detectors. Two-stage detectors, such as the faster region-based convolutional neural network (R-CNN) (Gao et al., 2020; Cui et al., 2024), are generally more accurate but are unsuitable for real-time applications due to slower processing. In contrast, one-stage detectors, such as you only look once (YOLO) model, have gained popularity owing to their speed and robustness, particularly in imbalanced datasets. For instance, YOLO outperformed the Fast R-CNN in identifying minority-class traits in (Solimani et al., 2024). Therefore, one-stage detectors are typically preferable in applications where high, real-time identification accuracy is necessary.

Among DL models, YOLO has gained significant traction in agriculture due to its real-time detection capability and high accuracy (Redmon et al., 2016). Researchers favor the YOLO series owing to its efficient architecture, strong feature extraction, and continuous updates (Zhang et al., 2025b). The YOLO framework divides an image into grids and sets bounding boxes and class probabilities for prediction in each region, making it suitable for various tasks (e.g., monitoring, sensing, automation, and robotic control in agriculture). Recent agricultural applications have adapted YOLO to specific tasks. For example, Pineapple-YOLO, based on YOLOv7-tiny, achieved 89.7% mAP@0.5 and 110 FPS, optimizing speed and moderate scene complexity (Li et al., 2025b). Similarly, YOLO-RGBDtea enhanced YOLOv7 with RGB-D fusion, spatial attention, and self-attention, achieving 91.12% mAP@0.5 and a 0.912 F1-score under challenging lighting and occlusion conditions (Wu et al., 2024). Despite these improvements, YOLOv7-based models still encounter trade-offs between the detection detail, computational efficiency, and robustness in high-density or embedded environments.

The YOLOv8-based architecture offers significant improvements in detection precision and deployment flexibility for agricultural applications. Faster-YOLO-AP, built on the YOLOv8 backbone, achieved an 84.12% mAP@0.50:0.95, making it ideal for edge deployment without sacrificing accuracy (Liu et al., 2024). The YOLOv8x model has also been applied to uncrewed aerial vehicle-based stem segmentation in dense potato canopies, applying RGB-DSM fusion and high-resolution sampling to achieve an F1-score of 0.86 and an IoU of 0.83 for high-throughput phenotyping (Jiang et al., 2024). Comparative studies have demonstrated that YOLOv8 outperforms YOLOv5 and YOLOv9 on tree detection tasks involving cherries and apples (Akdoğan et al., 2025). Recent studies have adapted YOLOv8 for fruit detection under field conditions (He et al., 2025b). Global–local vision configurations that combine a fixed camera with an eye-in-hand camera have also been used in agricultural robots to combine wide-area observation with close-range target sensing (He et al., 2025a).

### 2.2 Harvest ordering

Researchers have extensively studied complementary perception tasks (e.g., yield estimation (Koirala et al., 2019; Liu et al., 2025), collision avoidance (Suo et al., 2021), harvest date prediction (Lin et al., 2025), and quality inspection). Nevertheless, the problem of prioritizing which fruits to harvest remains underexplored, particularly in environments with occlusions, dense foliage, or incomplete sensory data (Li et al., 2025c; Zhu et al., 2025). This lack of sequence-level reasoning under uncertainty remains a primary limitation in deploying autonomous harvesting systems. Addressing this gap requires integrated methods that can adaptively determine the harvesting order while considering the environmental complexity and sensor uncertainty.

One of the fundamental strategies for planning harvest sequences encompasses prioritizing fruit based on height, assuming that higher fruit is more accessible and less likely to be occluded (Bac et al., 2017). More advanced frameworks have extended this idea by jointly planning sensing and harvesting actions in partially observable environments. These methods treat the problem as a TSP, allowing for dynamic updates to the harvest sequence as new pieces of fruit are detected during execution (Kurtser and Edan, 2020). Although effective in theory, these approaches are susceptible to perception errors that commonly occur under real-time conditions in occluded and cluttered agricultural fields.

Recent studies have adopted learning-based and heuristic techniques to improve robustness and planning quality. For example, a model based on the 3D-CNN was employed to generate a position-pose fruit map, creating a heuristic cost map supporting rapid target selection and optimal path computation (Zhang et al., 2023). Another approach applies density-based clustering to group nearby fruit and employs dynamic programming to determine the optimal picking order in each cluster (Dai et al., 2024). This method reduces the motion cost by minimizing inter-fruit travel distances. However, these techniques often rely on ideal visibility and assume a structured environment without occlusion, limiting their reliability in realistic field conditions with prevalent leaves, stems, and variable illumination. More recent work has also addressed sequential picking in dense crop environments by combining three-dimensional target information with robot path planning (Fan et al., 2026).

## 3. Fruit Perception System

### 3.1 Fruit detection

The initial step in robotic harvesting involves detecting and locating fruit in the canopy. This task necessitates a real-time recognition system with high precision and robustness in agricultural environments. The YOLOv8 model introduces an anchorless approach, eliminating the need for predefined anchor boxes and simplifying the detection pipeline (Shen et al., 2025). Moreover, YOLOv8 provides five scaled versions, ranging from lightweight models for low-resource devices to high-performance models optimized for speed and accuracy: YOLOv8n (nano), YOLOv8s (small), YOLOv8m (medium), YOLOv8l (large), and YOLOv8x (extra-large), differing based on the network depth and width. (Zhang et al., 2025a) and (Li et al., 2025a) reported that YOLOv8n is the least complex and fastest model among the compared YOLOv8 models, characterized by a smaller parameter size and lower computational demand. These attributes make YOLOv8n suitable for automated agricultural robotics, where a high response speed and real-time inference are critical to ensuring timely, efficient perception in dynamic farming environments. Therefore, YOLOv8n was selected as the fruit detector in this study, and the detected fruit positions were subsequently used in the localization procedure described in Section 3.2.

The YOLOv8n model comprises the input, backbone, neck, and output. The input stage applies image resizing and data augmentation, including mosaic augmentation and padding. Because YOLOv8 adopts an anchor-free detection head, predefined anchor-box computation is not required. The backbone extracts features and transforms images of diverse sizes into fixed-size feature vectors, enhancing the adaptability of the model to multiple target sizes. In addition, the backbone uses the C2f module instead of the traditional YOLO neck architecture, which is a faster and lighter implementation of the cross-stage partial (CSP) bottleneck. The CSP bottleneck splits the feature map into two parts, processing one part via a series of convolutional layers, and concatenating both parts. The C2f module is faster and lighter because it uses two convolutional layers instead of three, optimizing the feature map processing for improved performance. The neck integrates the feature pyramid network and path aggregation network to achieve top-down and bottom-up feature fusion, strengthening the ability of the model to integrate features. The output component applies enhanced feature computation to deliver confidence scores and location information for each target. Thus, the accuracy of the classification and regression tasks is improved, thus supporting precise object detection and classification.

### 3.2 Fruit localization

Accurate fruit localization is critical for autonomous harvesting robots because it directly influences grasping precision and task efficiency (Jiang et al., 2025). The global–local RGB-D sensing system is configured according to the distinct information requirements of harvest-order optimization and robot execution. A fixed global RGB-D camera observes multiple fruits within the harvesting workspace and provides the three-dimensional fruit positions used for harvest-order optimization. In contrast, the local RGB-D camera is mounted on the end-effector and provides close-range target observation during robot execution. The robot-base frame is used as the common reference frame for relating the global camera, end-effector, and local camera. The sensing configuration and coordinate frames are shown in Fig. 3.

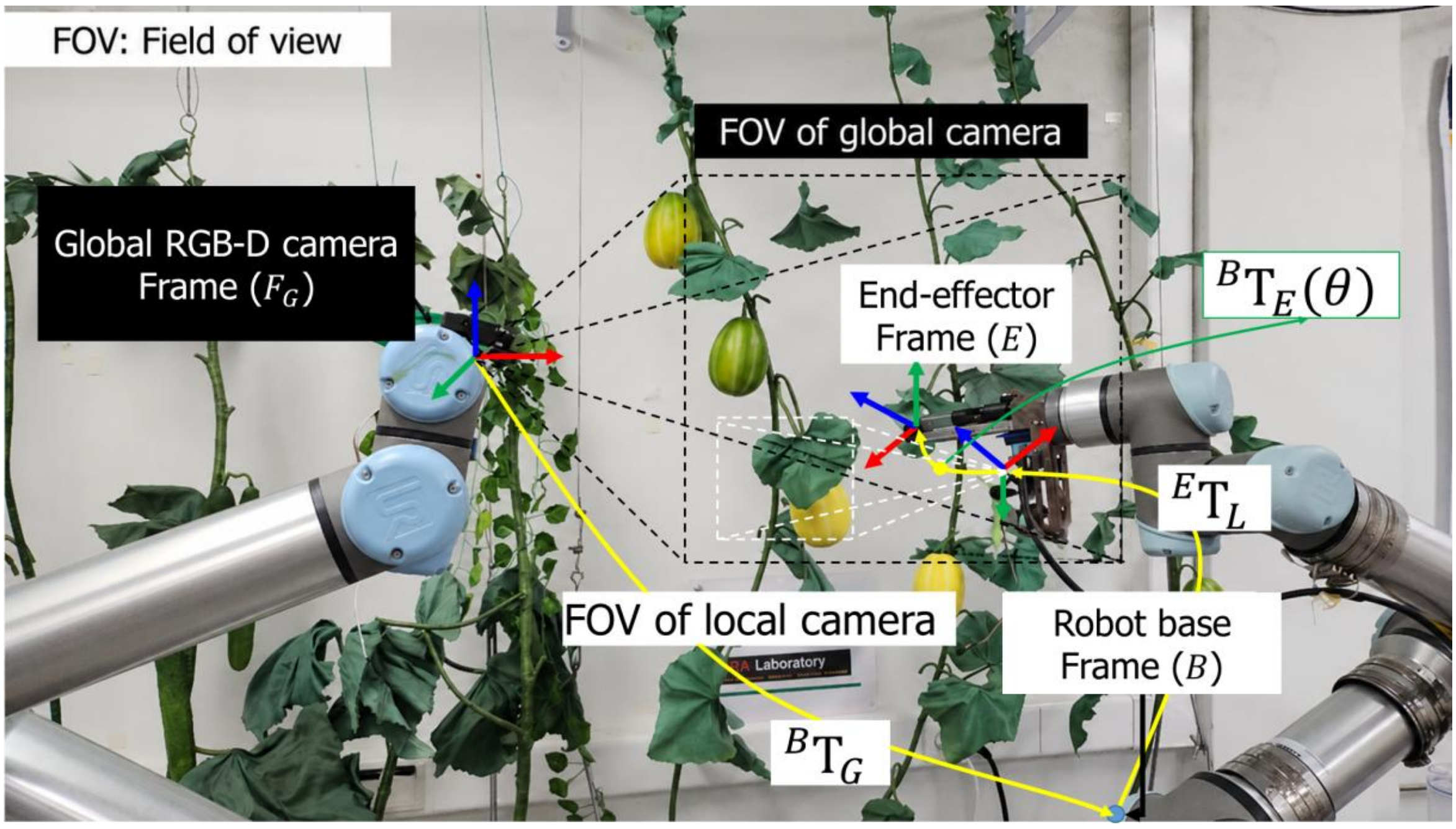

**Fig. 3** RGB-D sensing configuration and coordinate relationship between the global camera, robot base, end-effector, and local camera

For each fruit detected by the global RGB-D camera, the center of the detected region is associated with the corresponding pixel in the aligned depth image. Let $(u_i, v_i)$ denote the image coordinate of fruit $i$, and let $z_i^G$ denote the corresponding depth. Using the intrinsic parameters of the global camera, the fruit position in the global-camera frame $F_G$ is expressed in homogeneous coordinates as

$$ {}^{G}\tilde{\mathrm{p}}_i = \begin{bmatrix} (u_i - c_x) z_i^G / f_x \\ (v_i - c_y) z_i^G / f_y \\ z_i^G \\ 1 \end{bmatrix}, \tag{1} $$

where $f_x$ and $f_y$ are the focal lengths and $c_x$ and $c_y$ are the principal-point coordinates. The spatial relationship between the global camera and the robot base is described by a fixed rigid-body transformation, with rotation and translation determined using an ArUco marker as a spatial reference. The corresponding homogeneous transformation is expressed as

$$ {}^{B}\mathrm{T}_{G} = \begin{bmatrix} {}^{B}\mathrm{R}_{G} & {}^{B}\mathrm{t}_{G} \\ 0^{T} & 1 \end{bmatrix}. \tag{2} $$

where ${}^{B}\mathrm{R}_{G}$ and ${}^{B}\mathrm{t}_{G}$ denote the rotation and translation from the global-camera frame to the robot-base frame. The deprojected fruit position is transformed to the robot-base frame as

$$ {}^{B}\tilde{\mathrm{p}}_{i} = {}^{B}\mathrm{T}_{G}\,{}^{G}\tilde{\mathrm{p}}_{i}. \tag{3} $$

Equivalently, the three-dimensional fruit position used by the harvesting framework is expressed as:

$$ \mathrm{p}_{i} = {}^{B}\mathrm{R}_{G} \begin{bmatrix} (u_i - c_x) z_i^G / f_x \\ (v_i - c_y) z_i^G / f_y \\ z_i^G \end{bmatrix} + {}^{B}\mathrm{t}_{G}, \tag{4} $$

where $\mathrm{p}_{i} = [x_i, y_i, z_i]^T$ denotes the position of fruit $i$ in the robot-base frame. These positions are used for harvest-order optimization and waypoint generation.

Because the local RGB-D camera is rigidly mounted on the end-effector, its pose relative to the robot base changes with the manipulator configuration. Let ${}^{E}\mathrm{T}_{L}$ denote the fixed camera mounting transformation from the local-camera frame $F_L$ to the end-effector frame, and let ${}^{B}\mathrm{T}_{E}(\theta)$ denote the end-effector pose with respect to the robot-base frame, where $\theta$ is the vector of manipulator joint positions. Composing these transformations gives the local-camera pose relative to the robot base:

$$ {}^{B}\mathrm{T}_{L}(\theta) = {}^{B}\mathrm{T}_{E}(\theta)\,{}^{E}\mathrm{T}_{L}. \tag{5} $$

For a target position expressed in homogeneous coordinates ${}^{L}\tilde{\mathrm{p}}_{i}$ in the local-camera frame, the corresponding position in the robot-base frame is obtained as

$$ {}^{B}\tilde{\mathrm{p}}_{i}^{L} = {}^{B}\mathrm{T}_{L}(\theta)\,{}^{L}\tilde{\mathrm{p}}_{i} = {}^{B}\mathrm{T}_{E}(\theta)\,{}^{E}\mathrm{T}_{L}\,{}^{L}\tilde{\mathrm{p}}_{i}. \tag{6} $$

Here, ${}^{B}\tilde{\mathrm{p}}_i^{L}$ denotes the local-camera measurement expressed in the robot-base frame; the right superscript $L$ identifies its measurement source. The transformation from the local-camera frame to the global-camera frame follows from their common robot-base reference:

$$ {}^{G}\mathrm{T}_L(\theta) = ({}^{B}\mathrm{T}_G)^{-1}\,{}^{B}\mathrm{T}_L(\theta) = ({}^{B}\mathrm{T}_G)^{-1}\,{}^{B}\mathrm{T}_E(\theta)\,{}^{E}\mathrm{T}_L. \tag{7} $$

The global RGB-D camera provides the fruit coordinates used to determine the harvest order and generate the waypoint motion. During execution, the local RGB-D camera provides close-range observation of the selected target. The fruit coordinates used for harvest-order planning are retained from the global-camera measurements throughout execution.

## 4. Harvesting Ordering System

### 4.1 Static genetic algorithm-based path planning

The harvesting order can be determined based on several factors, such as crop maturity, quality, accessibility, harvesting difficulty, and work efficiency. In robotic harvesting, the spatial relationship among the target fruit is particularly important because repeated manipulator motion between spatially distributed targets directly affects the efficiency of sequential harvesting. Therefore, this study formulates the harvesting-order problem based on the three-dimensional spatial distribution of the recognized fruit. The notation used in this paper is summarized in Table 1.

The detected fruit set is represented as an undirected weighted graph $G = (V, E)$, where each vertex $v_i \in V$ corresponds to a target fruit and each edge $(i, j) \in E$ represents a possible transition between two fruit positions. Let the position of fruit $i$ in the robot base frame be expressed as $\mathrm{p}_i = [x_i, y_i, z_i]^T$. The transition cost between two target fruit is defined by the three-dimensional Euclidean distance:

$$ d_{i,j} = \|\mathrm{p}_j - \mathrm{p}_i\|_2 = \sqrt{(x_j - x_i)^2 + (y_j - y_i)^2 + (z_j - z_i)^2}. \tag{8} $$

**Table 1.** Notation used in this paper

| Symbol | Description |
|---|---|
| $N$ | Number of detected fruit |
| $p_i$ | Position of fruit $i$ in the robot-base frame |
| $d_{i,j}$ | Euclidean transition cost between fruits $i$ and $j$ |
| $S$ | Candidate harvesting sequence |
| $J_{ord}(S)$ | Harvest-ordering cost of sequence $S$ |
| $S^*$ | Optimized harvesting sequence |
| $\mathrm{w}_i^S$ | Static waypoint associated with fruit $i$ |
| $\delta_s$ | Static waypoint-offset magnitude |
| $\mathrm{r}_s$ | Static waypoint-offset direction |
| $R_k$ | Remaining-target set at harvesting step $k$ |
| $\mathrm{c}_k$ | Centroid of the remaining-target set at harvesting step $k$ |
| $\mathrm{w}_k^d$ | Dynamic waypoint at harvesting step $k$ |
| $\delta_d$ | Dynamic waypoint-offset magnitude |
| $\mathrm{r}_k$ | Dynamic waypoint-offset direction at harvesting step $k$ |
| $J_{traj}$ | TCP trajectory length |

By incorporating displacement along all three axes, including depth, this distance enables the ordering cost to represent the three-dimensional fruit distribution. To search for a sequence using this cost, the GA represents each detected fruit as a gene and each candidate sequence as an arrangement of these genes. Selection, crossover, and mutation are then applied iteratively to search the sequence space. As illustrated in Fig. 4, the resulting sequence determines the target order, while the waypoint strategy defines the intermediate robot motion needed to execute it.

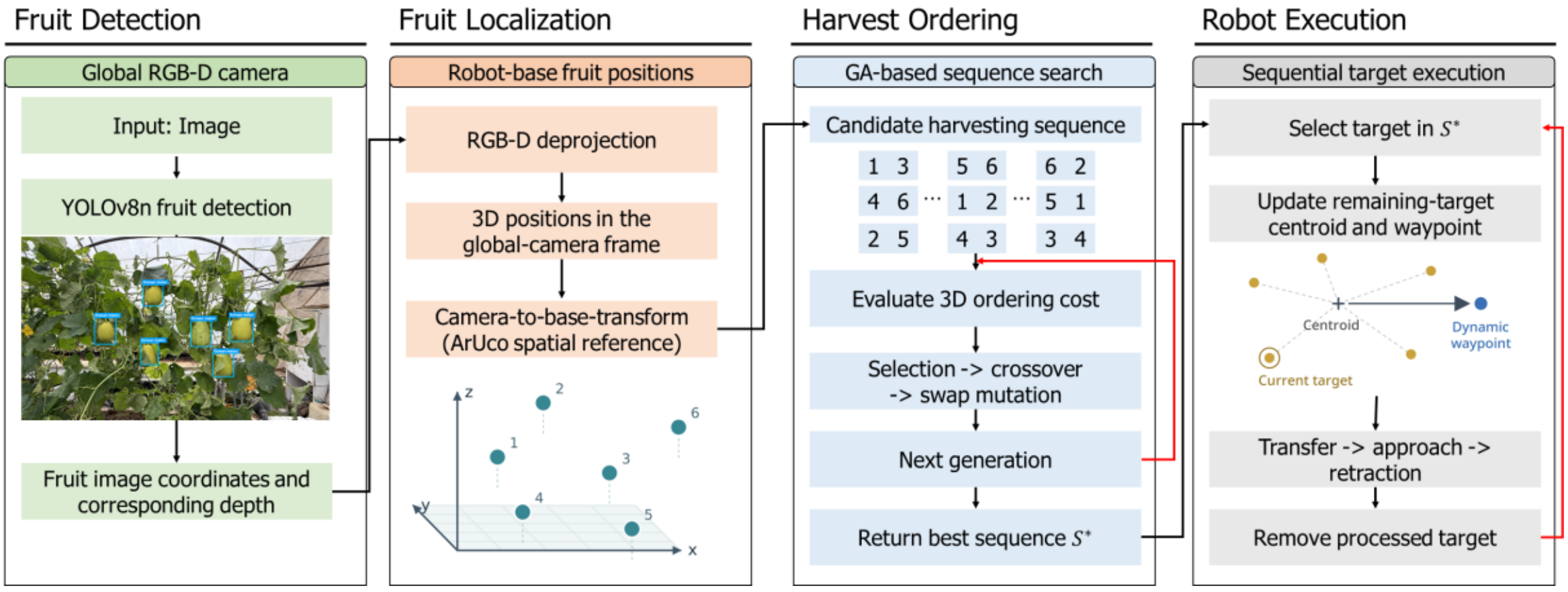

**Fig. 4** Flow chart of the harvesting robot system.

### 4.1.1 Generation of offset waypoint pairs

To distinguish the transfer motion between targets from the final target approach, an intermediate waypoint is assigned to each detected fruit. For a fruit position pi, the corresponding static waypoint is defined as

$$\mathrm{w}_i^s = \mathrm{p}_i + \delta_s \mathrm{r}_s, \tag{9}$$

where $\delta_s$ denotes the waypoint-offset magnitude and $\mathrm{r}_s$ is a unit vector defining the waypoint-offset direction in the robot-base frame, with $\|\mathrm{r}_s\|_2 = 1$. Thus, the static waypoint maintains a fixed geometric relationship with its corresponding fruit. Each fruit is consequently represented by an approach–target waypoint pair,

$$\mathrm{W}_i^s = (\mathrm{w}_i^s, \mathrm{p}_i). \tag{10}$$

In the implemented system, the waypoint-offset direction is aligned with the positive $x$-axis of the robot-base frame. Let $\mathrm{e}_x$ denote the unit vector along this axis, then $\mathrm{r}_s = \mathrm{e}_x$. The offset magnitude $\delta_s$ is treated as a tunable parameter of the waypoint-generation procedure. The setting used in the physical experiments and its geometric influence are examined in Section 5.3.1 using the recorded target configurations and a separate environment analysis of travel distance and end-effector interference. Once the fruit position and waypoint parameters are specified, the static waypoint remains fixed throughout the corresponding execution condition.

### 4.1.2 Genetic algorithm-based harvest sequence optimization

For N detected fruit, a candidate harvesting sequence is represented by the permutation

$$S = (\pi_1, \pi_2, \dots, \pi_N). \quad (11)$$

where $\pi_k$ denotes the fruit index assigned to the $k$th harvesting step. Using the transition cost defined in Eq. (8), the sequence-level objective is expressed as

$$J_{ord}(S) = \sum_{k=1}^{N-1} d_{\pi_k,\pi_{k+1}} . \quad (12)$$

An additional return term after the final target is not included in $J_{ord}$ because return and transfer motions depend on the robotic execution strategy and are treated separately from the harvest-order optimization. Therefore, the optimal harvesting sequence is defined as

$$S^* = \arg\min_S J_{ord}(S) . \quad (13)$$

The GA searches the permutation space by evolving a population of candidate harvesting sequences. Each chromosome contains every detected fruit exactly once. For each generation, the ordering cost of every chromosome is evaluated, favorable candidates are retained through selection, and crossover and mutation are applied to produce valid candidate sequences for the next generation. The GA configuration is defined by the population size $M$, the number of generations $N_{gen}$, the retained-population size $M_{keep}$, and the number of swap mutations $N_{mut}$.

The GA parameters are treated as configuration parameters of the optimization procedure. The values used for the reported experiments are specified in the experimental setup. After the optimized sequence $S^*$ is obtained, Algorithm 1 arranges the target-specific static waypoint pairs according to this sequence.

**Algorithm 1: Genetic Algorithm-based Harvest-Order Optimization with Static Waypoint Generation**

**Input:** Fruit positions $\{\mathrm{p}_i\}_{i=1}^{N}$, static waypoint parameters $(\delta_s, \mathrm{r}_s)$, and GA parameters $(M, N_{gen}, M_{keep}, N_{mut})$
**Output:** Optimized harvesting sequence $S^*$ and ordered static waypoint pairs $\mathrm{W}_{S^*}^{S}$
Construct weighted fruit graph $G = (V, E)$
**for** each target pair $(i, j)$ **do**
    Compute transition cost $d_{i,j} = \|\mathrm{p}_j - \mathrm{p}_i\|_2$
**end for**
Initialize GA population $P^{(0)} = \{S_1^{(0)}, \dots, S_M^{(0)}\}$
**for** $g = 0$ to $N_{gen} - 1$ **do**
    **for** each sequence $S_m^{(g)} \in P^{(g)}$ **do**
        Compute cost $J_{ord}(S_m^{(g)})$
        Assign fitness according to $J_{ord}(S_m^{(g)})$
    **end for**
    Rank candidate sequence according to fitness
    Retain $M_{keep}$ favorable candidate sequences
    Select parent sequences from the retained population
    Apply crossover to generate valid permutations
    Apply $N_{mut}$ swap mutations while preserving valid permutations
    Construct the next population $P^{(g+1)}$
**end for**
Select $S^*$ as the sequence with the lowest ordering cost among all candidates evaluated during the GA search
Initialize ordered waypoint-pair set $W_{S^*}^{S} \leftarrow [\ ]$
**for** $k = 1$ to $N$ **do**
    Generate static waypoint $\mathrm{w}_{\pi_k^*}^{\mathrm{S}} = \mathrm{p}_{\pi_k^*} + \delta_s \mathrm{r}_s$
    Append $\left(\mathrm{w}_{\pi_k^*}^{\mathrm{S}}, \mathrm{p}_{\pi_k^*}\right)$ to $W_{S^*}^{S}$
**end for**
Return $S^*$ and $W_{S^*}^{S}$

Because the evaluated harvesting scenes contain a small number of target fruit, the GA solution can additionally be compared with the exact solution obtained through exhaustive enumeration. This comparison provides an independent assessment of the solution quality of the selected GA configuration. The corresponding analysis is presented in Section 5.3.1.

### 4.2 Dynamic waypoint augmented genetic-algorithms-based path planning

The static waypoint strategy defines an intermediate approach position relative to each individual fruit, whereas sequential execution progressively changes the distribution of the remaining targets. The dynamic waypoint strategy accounts for these changes by recalculating the intermediate position from the remaining-target set after each harvesting step. The GA sequence, determined from the complete three-dimensional fruit distribution, is retained throughout execution; the waypoint update adapts the intermediate robot motion to the remaining-target distribution.

#### 4.2.1 Dynamic waypoint generation for adaptive execution

The dynamic waypoint is constructed from the centroid of the remaining-target set, which provides a spatial reference for inter-target transfer. To define this set at each harvesting step, let the fixed optimized sequence be represented by

$$S^* = (\pi_1^*, \pi_2^*, \dots, \pi_N^*). \tag{14}$$

At harvesting step $k$, the current target and all targets scheduled for subsequent harvesting define the remaining-target set

$$R_k = \left\{ \mathrm{p}_{\pi_j^*} \mid j = k, \dots, N \right\}. \tag{15}$$

The centroid of this set provides the spatial reference for the intermediate waypoint:

$$\mathrm{c}_k = \frac{1}{|R_k|} \sum_{\mathrm{p}_j \in R_k} \mathrm{p}_j . \tag{16}$$

**Algorithm 2: Dynamic Waypoint-Augmented Genetic Algorithm-based Harvest Planning**

**Input:** Fruit positions $\{\mathrm{p}_i\}_{i=1}^{N}$, GA parameters $(M, N_{gen}, M_{keep}, N_{mut})$, and dynamic waypoint parameters $(\delta_d, \mathrm{r}_k)$

**Output:** Optimized harvest sequence $S^*$ and dynamically refined execution trajectory $J_{S^*}^{d}$

Obtain the harvesting sequence $S^* = (\pi_1^*, \pi_2^*, \dots, \pi_N^*)$ using the GA optimization stage of Algorithm 1

Initialize $J_{S^*}^{d} \leftarrow [\ ]$

Initialize $R_1 = \{\mathrm{p}_{\pi_1^*}, \dots, \mathrm{p}_{\pi_N^*}\}$

**for** $k = 1$ to $N$ **do**

  Select current target $\mathrm{p}_{\pi_k^*}$ according to $S^*$

  Update remaining-target set $R_k = \left\{\mathrm{p}_{\pi_j^*} \mid j = k, \dots, N\right\}$

  Compute centroid $\mathrm{c}_k = |R_k|^{-1} \sum_{\mathrm{p}_j \in R_k} \mathrm{p}_j$

  Determine waypoint-offset direction $\mathrm{r}_k$

  Generate dynamic waypoint $\mathrm{w}_k^d = \mathrm{c}_k + \delta_d \mathrm{r}_k$

  Plan transfer motion toward $\mathrm{w}_k^d$

  Plan target-approach motion from $\mathrm{w}_k^d$ to $\mathrm{p}_{\pi_k^*}$

  Execute the planned approach motion

  Generate retreat motion from $\mathrm{p}_{\pi_k^*}$ toward $\mathrm{w}_k^d$

  Concatenate the planned motion segments to $J_{S^*}^{d}$

  Remove $\mathrm{p}_{\pi_k^*}$ from the active target set

**end for**

Evaluate the resulting execution trajectory $J_{traj}(J_{S^*}^{d})$

Return $S^*$ and $J_{S^*}^{d}$

The dynamic waypoint is then generated by offsetting this centroid along a prescribed direction:

$$\mathrm{w}_k^d = \mathrm{c}_k + \delta_d \mathrm{r}_k \tag{17}$$

where $\delta_d$ denotes the dynamic waypoint-offset magnitude and $\mathrm{r}_k$ is a unit vector defining the waypoint-offset direction at harvesting step $k$, with $\|\mathrm{r}_k\|_2 = 1$. Unlike the static waypoint in Eq. (9), whose reference position is associated with an individual fruit, the dynamic waypoint is determined from the spatial distribution of the remaining targets.

In the implemented system, the offset direction is aligned with the positive $x$-axis of the robot-base frame. Let $\mathrm{e_x}$ denote the unit vector along this axis; then $\mathrm{r}_k = \mathrm{e}_x$. The offset magnitude $\delta_d$ is treated as a tunable parameter of the dynamic waypoint generation procedure. Section 5.3.1 presents the evaluated offset values and the setting used in the physical experiments. The recorded-coordinate analysis quantifies the resulting path length and waypoint–target distance, while the separate environment analysis examines the balance between travel distance and modeled end-effector interference.

During execution, the remaining-target set $R_k$ and its centroid $\mathrm{c}_k$ are recalculated after each harvesting step. Consequently, $\mathrm{w}_k^d$ changes according to the updated target distribution while the optimized harvesting sequence $S^*$ remains unchanged. This online update of the intermediate waypoint is referred to as dynamic waypoint refinement.

### 4.2.2 Genetic algorithm-based planning with dynamic waypoint refinement

Algorithm 2 applies dynamic waypoint refinement while retaining the GA-based sequence $S^*$. At harvesting step $k$, the robot selects $\mathrm{p}_{\pi_k^*}$ from this sequence, constructs the remaining-target set $R_k$, and computes its centroid $\mathrm{c}_k$. The waypoint $\mathrm{w}_k^d$ generated by Eq. (17) then provides the intermediate transition position before the robot approaches the current target. After the target operation, the processed fruit is removed from the active set, so the next step uses an updated centroid and waypoint. This process continues until all targets in $S^*$ have been processed. Let $J = \{\mathrm{q}_1, \mathrm{q}_2, \dots, \mathrm{q}_L\}$ denote the Cartesian trajectory generated during robotic execution, where $\mathrm{q}_l$ represents a sampled TCP position along the planned or executed motion. The geometric trajectory length is defined as

$$J_{traj}(J) = \sum_{l=1}^{L-1} \|\mathrm{q}_{l+1} - \mathrm{q}_l\|_2. \tag{18}$$

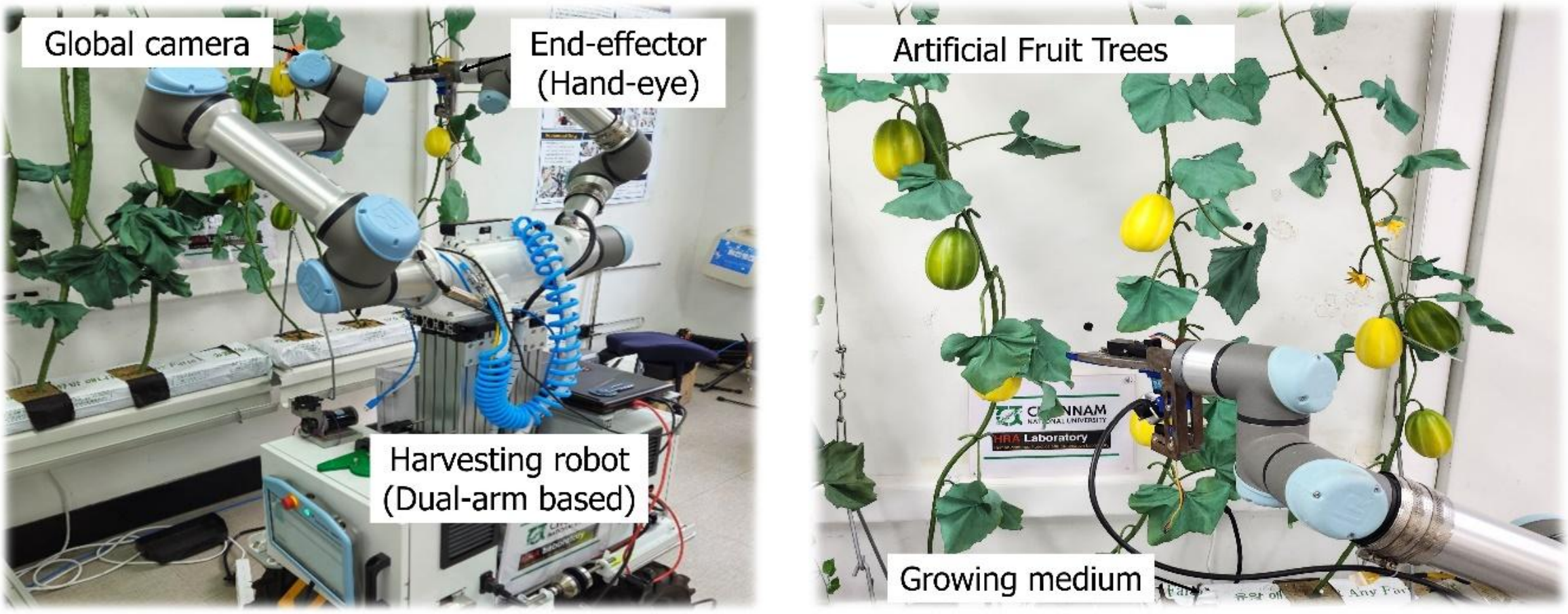

(a) (b)

**Fig. 5** Experimental setup: (a) hardware and (b) testbed was configured with a manipulator, RGB-D camera, and artificial fruits to reflect dense field conditions.

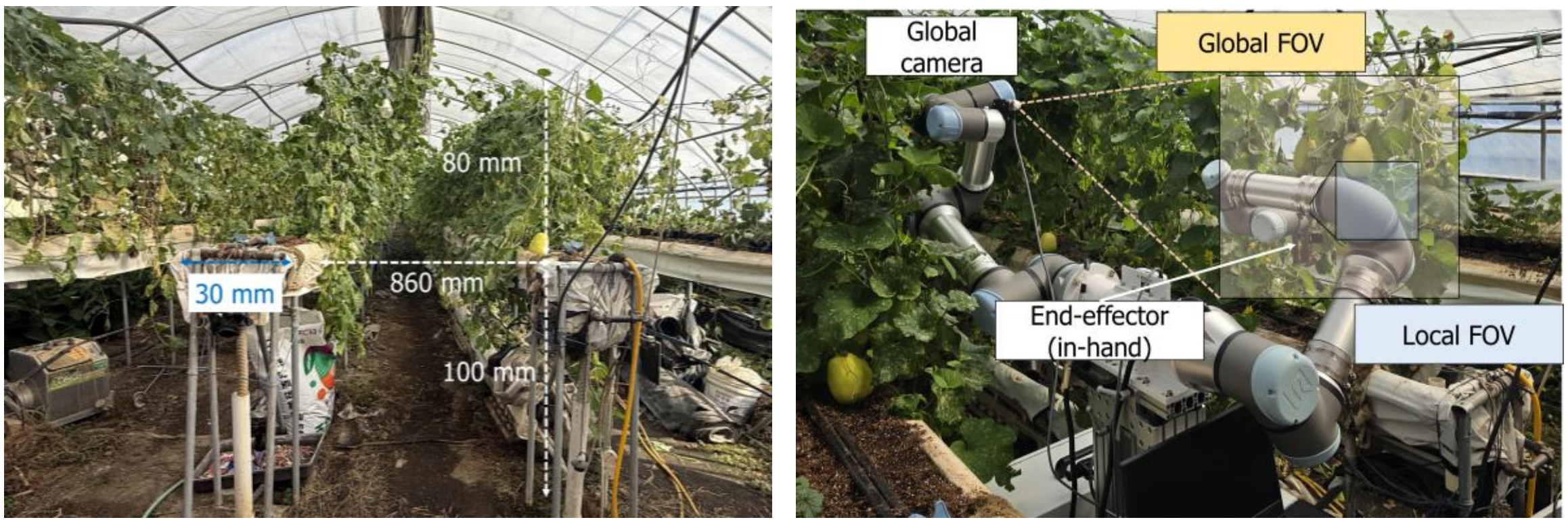

**Fig. 6** Field experimental environment in a greenhouse; overview of the harvesting workspace, showing the spatial arrangement of fruit and surrounding foliage under actual growing conditions.

This trajectory measure describes the manipulator motion generated by the waypoint strategy, whereas $J_{ord}(S)$ in Eq. (12) is the objective used to determine the harvesting order. Dynamic waypoint refinement therefore changes the motion associated with $J_{traj}(J)$ while retaining the sequence $S^*$ obtained from the ordering objective.

## 5. Experiment

### 5.1 Experimental setup

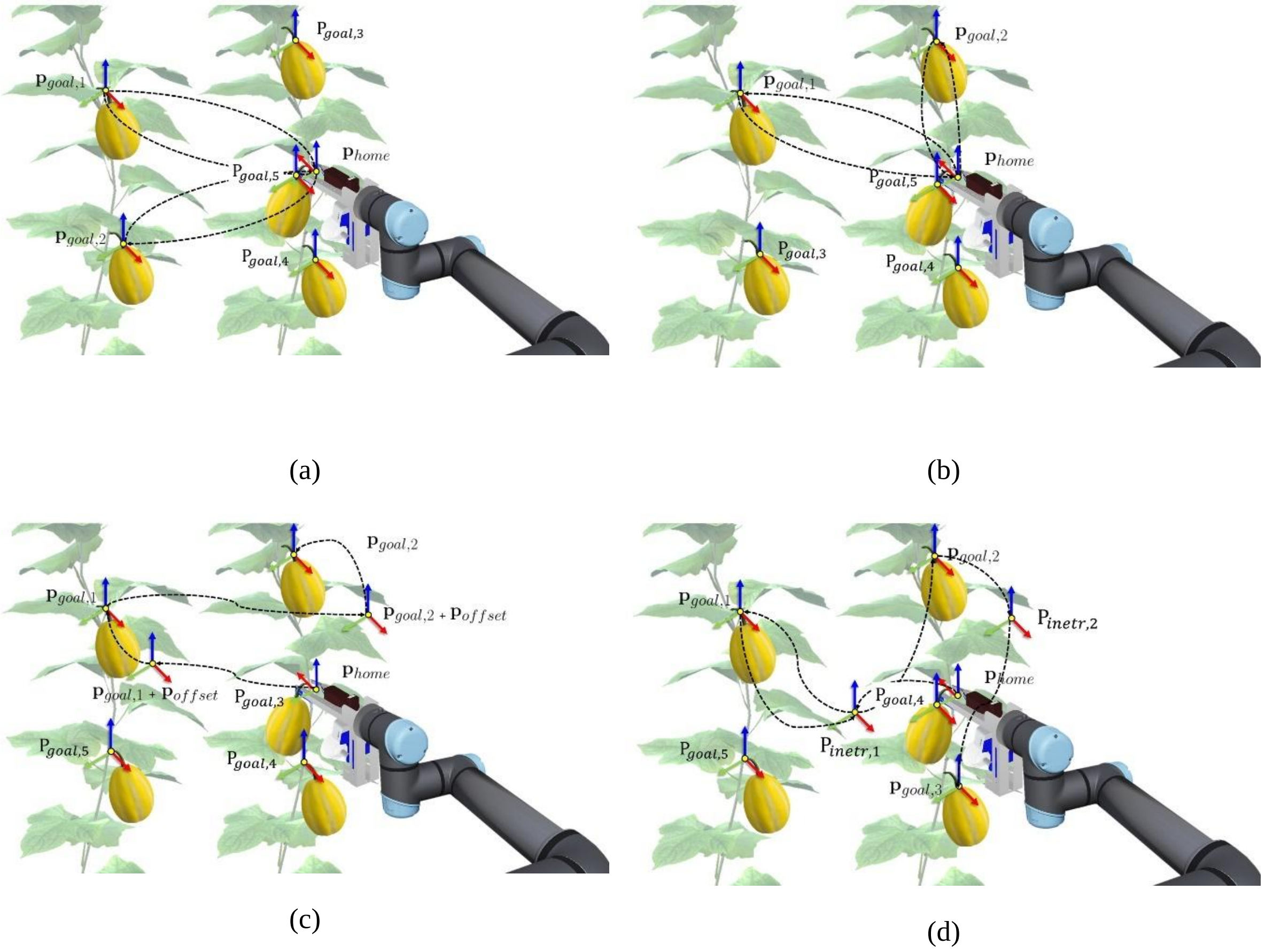

**Fig. 7** Experimental scenarios: (a) case 1, (b) case 2, (c) case 3, and (d) case 4.

The experimental setup supported fruit detection, harvest-order planning, and robot execution in the laboratory and greenhouse. The hardware, perception dataset, and parameter settings are described below.

### 5.1.1. Experimental environments and hardware

The robotic harvesting system was evaluated in a controlled laboratory testbed and a greenhouse in Nonsan, Republic of Korea. The laboratory testbed shown in Fig. 5 used artificial fruit, whereas the greenhouse environment shown in Fig. 6 contained naturally occurring variation in fruit count and spatial arrangement. In both environments, the experiments used the right manipulator of the dual-arm platform. A global RGB-D camera provided the fruit positions for harvest-order planning, while a local RGB-D camera mounted on the manipulator provided close-range observation during robot execution. The detector-training hardware consisted of an Intel Core i7-10750H CPU, an NVIDIA GeForce RTX 2060 GPU, and 16 GB of RAM.

### 5.1.2. Dataset construction

The detection dataset comprised images of yellow Korean melons and their leaves acquired in the laboratory testbed using an iPhone 16 Pro camera. To reflect the side views captured by the robot camera, images were acquired from lateral viewpoints, with the viewing angle varied to provide perspective diversity. A total of 282 images were collected and annotated using Roboflow, with Korean melons and leaves defined as the two object classes. The annotated dataset was divided into 198 training images, 42 validation images, and 42 test images.

### 5.1.3. Parameter settings

The perception models were trained using the dataset described above. All models were trained for 100 epochs with an input resolution of 640 × 640 pixels. The batch size and initial learning rate were set to 16 and 0.001, respectively, following preliminary training trials. These settings defined the training conditions for the detector comparison. For harvest-order optimization, the GA configuration was held constant across the experiments, with a population size of $M = 200$, a generation limit of $N_{\text{gen}} = 300$, a retained population size of $M_{\text{keep}} = 100$, and $N_{\text{mut}} = 20$ swap mutations. The separate offline validation used the same configuration with reference-assisted early termination, as described in Section 5.3.1.

### 5.2 Experimental design

The laboratory condition provided a repeatable comparison of the harvesting strategies, while the field condition examined their execution under variation in target count and spatial distribution. Both conditions used the same four execution strategies, illustrated in Fig. 7 and defined as follows:

- Case 1: Random sequence with direct target motion and Home return
- Case 2: GA-based sequence with static waypoints and Home return

- Case 3: GA-based sequence with static waypoints and continuous execution
- Case 4: GA-based sequence with dynamic waypoints and continuous execution

Cases 2 and 3 assess the effect of repeated Home return, whereas Cases 3 and 4 compare waypoint formulations under continuous execution. Cases 1 and 2 assess the combined changes in ordering and approach structure. Continuous execution assumes separate fruit transfer, whose mechanism was outside the evaluation. Each case was tested over 20 laboratory trials with six artificial fruit arranged at different depths with partial occlusion. Field trials covered four-, five-, and six-fruit conditions, with five trials per case and condition, yielding 60 trials and 75 target operations per case. Fruit coordinates were recorded before each execution to document the target configuration used in each trial.

Motion efficiency was assessed using execution time and travel length, while field videos were reviewed for collision events and successful target approaches. Uninterrupted contact counted as one event, and renewed contact after separation counted as another. An approach was successful when the end-effector reached the intended target region without crop structures blocking a harvestable approach. Since contact could occur before successful target access, the two outcomes were recorded independently. Planning and sequence-completion indicators were also recorded to distinguish command execution from physical target access. Fruit detachment and post-harvest condition were not evaluated.

**Table 2.** Detection results

| Model | Precision (%) | Recall (%) | mAP@0.50 (%) | mAP@0.50:0.95 (%) | F1-score (%) |
|---|---|---|---|---|---|
| YOLOv8n | 94.2 | 96.9 | 97.8 | 96.6 | 95.6 |
| YOLOv8s | 89.0 | 94.8 | 93.2 | 88.1 | 91.8 |
| YOLOv8m | 82.7 | 88.5 | 85.8 | 81.8 | 85.5 |
| YOLOv8l | 81.5 | 86.0 | 84.6 | 79.8 | 83.7 |

mAP@0.50 denotes AP at an IoU threshold of 0.50; mAP@0.50:0.95 denotes mean AP over IoU thresholds from 0.50 to 0.95

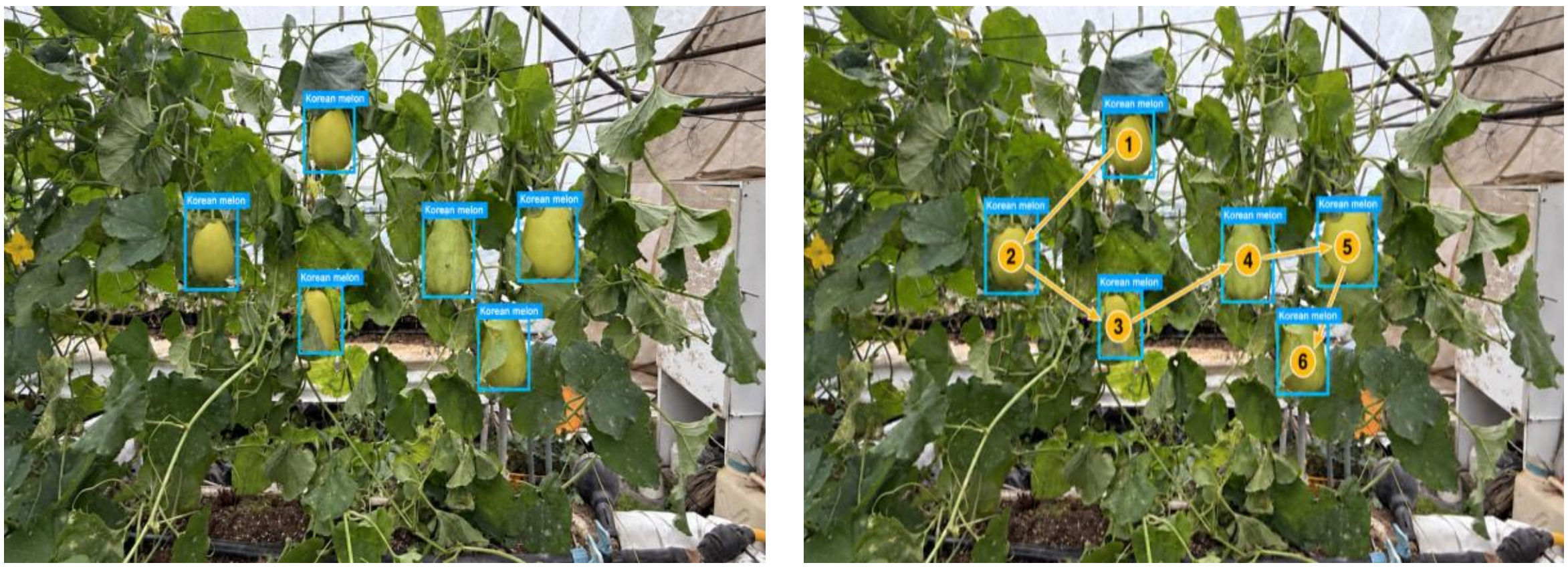

(a) (b)

**Fig. 8** Representative fruit detection and harvest-order outputs: (a) detected fruit and (b) the resulting GA-based harvest order.

**Table 3.** Offset-parameter analysis using recorded fruit coordinates and crop configurations.

| Offset (m) | Coordinate analysis | | | | Geometric analysis | | | |
|---|---|---|---|---|---|---|---|---|
| | Path length (m/target) | | Waypoint–target distance (m) | | Path length (m/target) | | Intersecting target steps (%) | |
| | St | Dy | St | Dy | St | Dy | St | Dy |
| 0.03 | 0.156±0.029 | 0.216±0.052 | 0.03 | 0.091±0.023 | 0.233±0.023 | 0.367±0.046 | 36.2 | 41.6 |
| 0.05 | 0.196±0.029 | 0.238±0.048 | 0.05 | 0.101±0.021 | 0.273±0.023 | 0.382±0.046 | 26.7 | 34.7 |
| 0.06† | 0.216±0.029 | 0.250±0.046 | 0.06 | 0.108±0.020 | **0.293±0.023** | 0.391±0.046 | **22.2** | 31.8 |
| 0.09 | 0.276±0.029 | 0.292±0.041 | 0.09 | 0.129±0.017 | 0.353±0.023 | 0.422±0.045 | 16.2 | 23.8 |
| 0.10‡ | 0.296±0.029 | 0.307±0.039 | 0.10 | 0.136±0.017 | 0.373±0.023 | **0.434±0.044** | 15.1 | **22.2** |
| 0.15 | 0.396±0.029 | 0.390±0.032 | 0.15 | 0.177±0.013 | 0.473±0.023 | 0.501±0.042 | 12.4 | 15.8 |

St and Dy denote static and dynamic waypoints, respectively. Values are mean ± sample SD across configurations. Path length is normalized by target count before aggregation; dynamic waypoint-target distance is averaged within each configuration first. Static waypoint-target distance equals $\delta_s$. Paths include inter-waypoint transfer, approach, and retraction, excluding Home travel. Intersection percentages are calculated over 450 target steps for each strategy and offset under the reference condition. † and ‡ indicate the static and dynamic settings used in the physical experiments, respectively. Bold values identify the geometric path length and interference results at these settings.

### 5.3 Experimental results

Table 3 compares the detection performance of the evaluated YOLOv8 variants. YOLOv8n achieved the highest precision, recall, mAP, and F1-score and was selected to provide detections for RGB-D localization and harvest-order planning. Figure 8 illustrates the corresponding fruit detections and GA-based harvesting order.

#### 5.3.1. Parameter analysis

Waypoint offset sensitivity was evaluated at the values listed in Table 2, using the same 30 recorded coordinate sets from Cases 3 and 4 for both formulations, with equal representation of each fruit-count condition. Fruit coordinates, harvesting order, and continuous execution policy were fixed within each set. Larger offsets increased path length for both formulations. For static waypoints, the increase arose from longer approach and retraction distances, since the waypoint–target distance equals the prescribed offset. For dynamic waypoints, this distance also depended on the target's position relative to the remaining-target centroid.

Geometric interference was evaluated in fruit and foliage layouts, balanced across the fruit-count conditions. Both formulations used the same shortest open Euclidean order within each layout. The model represented fruit and leaves with simplified geometries and the end-effector with a spherical envelope, allowing intended contact with the current fruit; leaf deformation, manipulator-link collisions, and kinematic feasibility were not modeled. In the reference condition, with three leaves per fruit and a 20 mm tool radius, larger offsets reduced intersecting target steps but increased path length, and static waypoints produced fewer intersections at matched offsets. A post hoc score subtracted the normalized increase in mean total path length from the normalized reduction in intersecting steps, with equal weights over the tested range. The score was highest at 0.09 m for both formulations under the reference condition. Across the tested variations in foliage density and tool radius, the highest-scoring offset remained at 0.09 m for dynamic waypoints and varied between 0.06 and 0.09 m for static waypoints. These results characterize the geometric effects of offset variation, while the marked settings identify the configurations evaluated in the physical experiments.

GA solution quality was assessed against the exact minimum ordering cost obtained by exhaustive enumeration for each recorded coordinate set. During offline validation, a run terminated when its best cost was within $10^{-12}$m of the reference minimum or the generation limit was reached. The reported runtime therefore describes GA search with reference-assisted termination. With $S_{\mathrm{GA}}$ and $S_{\mathrm{exact}}$ denoting the GA and exact solutions, respectively, the optimality gap (OG) was defined as

$$\mathrm{OG} = \frac{J_{ord}(S_{\mathrm{GA}}) - J_{ord}(S_{\mathrm{exact}})}{J_{ord}(S_{\mathrm{exact}})} \times 100. \tag{19}$$

**Table 4.** Offline GA validation against exhaustive enumeration.

| Evaluation item | Result |
|---|---|
| Recorded target-coordinate sets | 60 |
| GA runs per target set | 30 |
| Total GA runs | 1,800 |
| Exact optimum attainment | 100.0% |
| Mean optimality gap | $1.67 \times 10^{-15}$ % |
| Maximum optimality gap | $1.89 \times 10^{-14}$ % |
| Mean computation time (ms/run) | 4.0 |

Attainment denotes an optimality gap no greater than $10^{-9}$%. Mean computation time averages the set-level mean runtimes and excludes exhaustive enumeration. Runs use the reference-assisted stopping criterion described in the text.

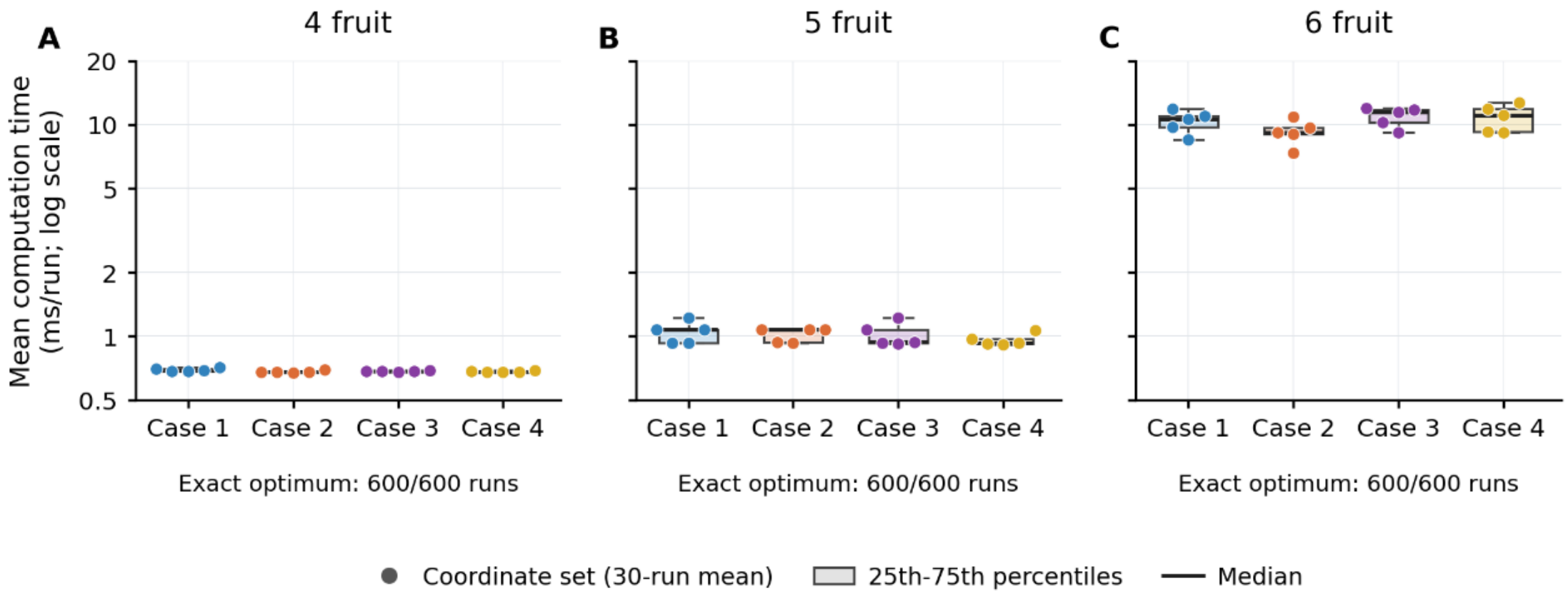

**Fig. 9** Offline GA computation time for (A) four-, (B) five-, and (C) six-fruit conditions. Each point represents the mean runtime for one coordinate set, with Case labels indicating its source. Boxes summarize the five set-level means in each group using the median and interquartile range; whiskers extend to observed values within 1.5 times the interquartile range. All observations are shown on a common logarithmic axis. Runtime definitions are given in Table 4.

All repeated runs attained the exact ordering optimum within numerical tolerance (Table 4), although the computation time varied across coordinate sets (Fig. 9). This agreement establishes the ordering quality for the evaluated inputs. The physical experiments then assess how the ordering and waypoint strategies translate into robot motion and target access

**Table 5.** Laboratory results of the harvesting strategies.

| Case | Harvesting time (s) | Collision (n) | Travel length (m) |
|---|---|---|---|
| Case 1 | 86.4±2.9 | 36.4±2.0 | 18.3±3.8 |
| Case 2 | 62.4±2.6 | 18.4±2.0 | 14.3±0.8 |
| Case 3 | 43.4±4.7 | 2.4±1.2 | 9.7±2.6 |
| Case 4 | 50.3±2.9 | 1.2±0.8 | 10.1±1.1 |

Values are mean ± sample SD across the 20 trials for each Case.

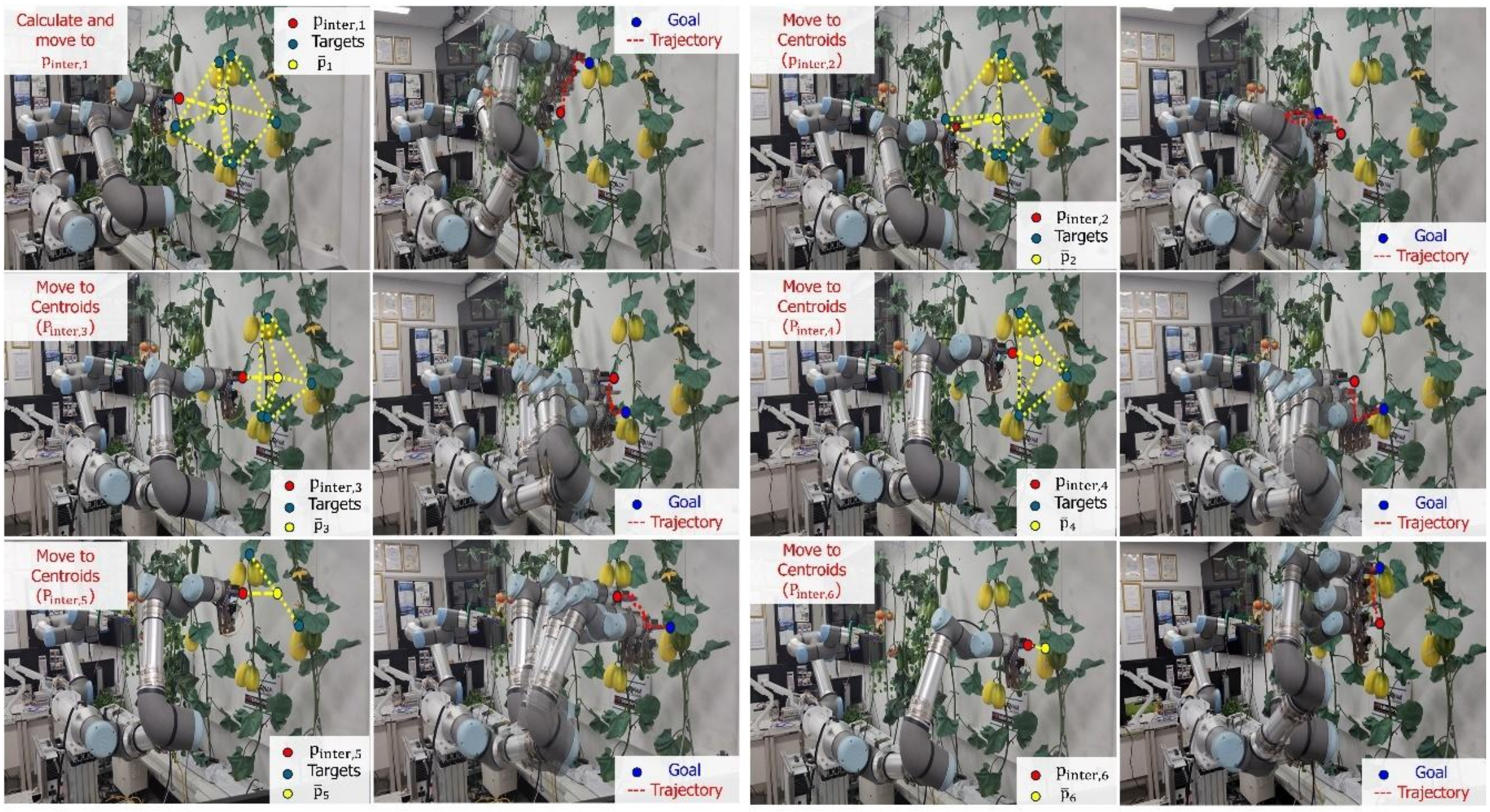

**Fig. 10** Laboratory execution with dynamic waypoints over six successive target steps. Each step shows intermediate waypoint motion and the final target approach; annotations identify the waypoint and remaining-target centroid.

### 5.3.2. Laboratory experiment

The laboratory comparison showed that reducing Home-return motion and changing waypoint placement affected different aspects of robot execution (Table 5). Case 2 reduced harvesting time, collision occurrence, and travel length relative to Case 1 through the combined use of GA-based ordering and static waypoints. Removing repeated Home return in Case 3 further reduced time and travel length. Under the same continuous execution policy, Case 4 achieved the lowest collision occurrence with dynamic waypoints, although time and travel length increased relative to Case 3. Figure 10 illustrates the intermediate waypoint motion and final target approach during dynamic-waypoint execution. The greenhouse experiments examined whether this pattern persisted among actual fruit and foliage.

### 5.3.3. Field experiment

In the greenhouse, Case 3 required less time and TCP travel, whereas Case 4 achieved fewer collisions and more successful target approaches (Table 6). Across the separately recorded configurations, Case 4 reduced

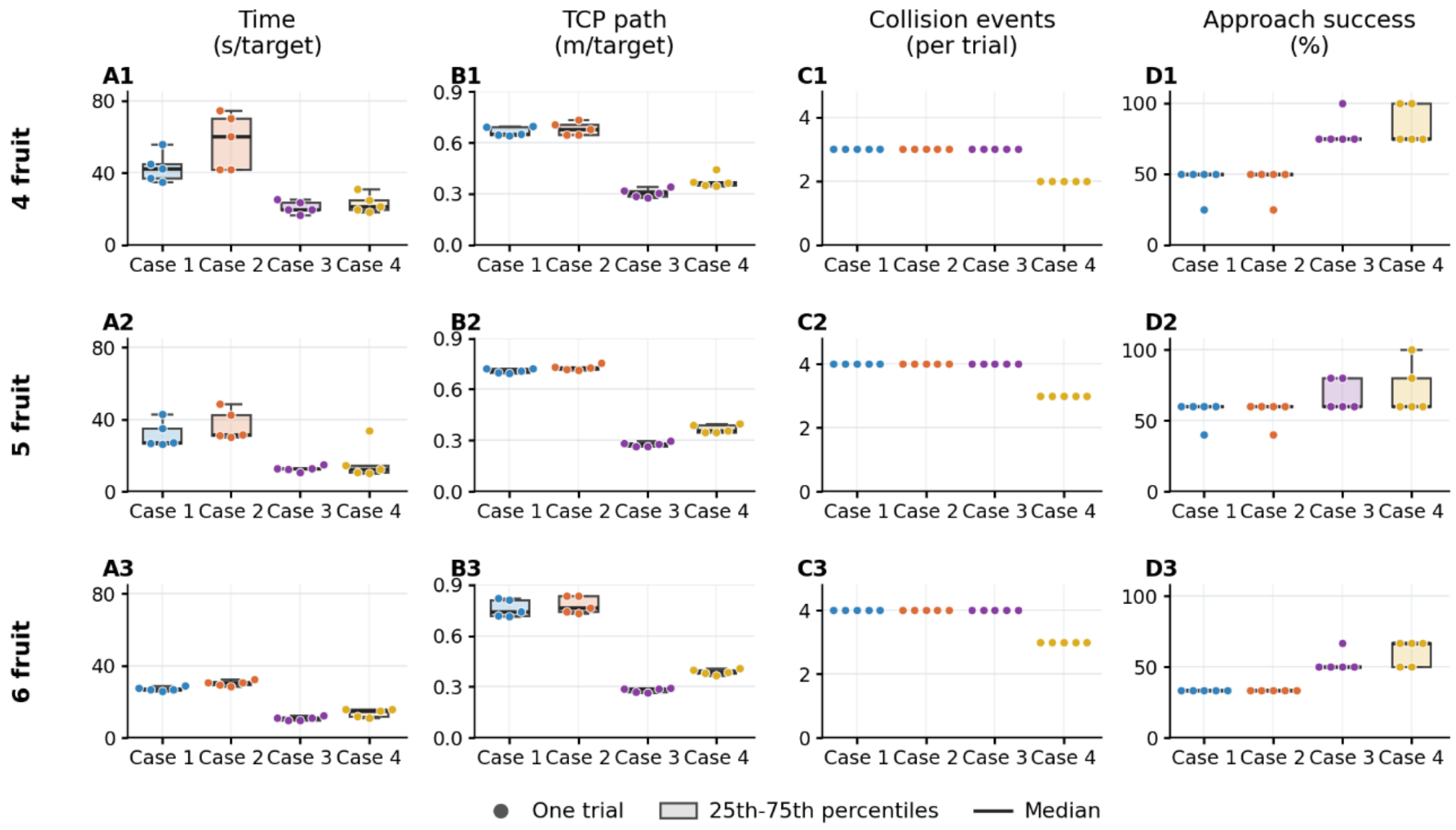

**Fig. 11** Field performance by Case and fruit count. Columns show (A1–A3) time per target, (B1–B3) TCP path per target, (C1–C3) collision events, and (D1–D3) approach success; rows correspond to four-, five-, and six-fruit conditions. Each point represents one trial. Boxes show the median and interquartile range, with whiskers extending to observations within 1.5 times the interquartile range. All observations are displayed; horizontal displacement separates overlapping points. Collision events are counted per trial, while Table 6 gives Case totals. Approach success follows the physical-access criterion in Section 5.2.

**Table 6.** Field motion efficiency and target accessibility under multiple fruit configurations

| Case | Motion efficiency | | Target accessibility | |
|---|---|---|---|---|
| | Time (s/target) | TCP path (m/target) | Collision events | Approach success n/N (%) |
| Case 1 | 33.92 ± 9.04 | 0.713 ± 0.051 | 55 | 33/75 (44.0) |
| Case 2 | 41.60 ± 15.37 | 0.731 ± 0.054 | 55 | 33/75 (44.0) |
| Case 3 | 14.77 ± 5.05 | 0.288 ± 0.021 | 55 | 49/75 (65.3) |
| Case 4 | 17.71 ± 7.29 | 0.377 ± 0.027 | 40 | 53/75 (70.7) |

Time and TCP path are mean ± sample SD across trials within each Case, after normalization by target count. Collision events are Case totals; approach success is the pooled number of successful approaches divided by total target attempts.

**Table 7.** Field motion-planning and sequence-completion indicators.

| Case | Planning success (%) | Target-level completion, n/N (%) | Trial-level completion, n/N (%) | Replanning n/N | Planning latency (s) |
|---|---|---|---|---|---|
| Case 1 | 100.0 | 75/75 (100.0) | 15/15 (100.0) | 0/75 | 0.0406 |
| Case 2 | 99.3 | 74/75 (98.7) | 14/15 (93.3) | 0/75 | 0.0265 |
| Case 3 | 100.0 | 75/75 (100.0) | 15/15 (100.0) | 7/75 | 0.0184 |
| Case 4 | 100.0 | 75/75 (100.0) | 15/15 (100.0) | 13/75 | 0.0178 |

Planning success is calculated over motion-planning attempts; completion refers to commanded target- or trial-level sequences. Replanning counts target sequences requiring at least one additional planning attempt. Latency is the unweighted mean of trial-level mean motion-planning times.

**Table 8.** Field execution results according to the number of observed fruit.

| Case | Time (s/target) | | | TCP path (m/target) | | |
|---|---|---|---|---|---|---|
| | 4 fruit | 5 fruit | 6 fruit | 4 fruit | 5 fruit | 6 fruit |
| Case 1 | 42.94 ± 8.26 | 31.61 ± 7.19 | 27.21 ± 1.11 | 0.667 ± 0.027 | 0.710 ± 0.013 | 0.761 ± 0.051 |
| Case 2 | 57.76 ± 15.52 | 36.72 ± 8.29 | 30.31 ± 1.43 | 0.684 ± 0.038 | 0.729 ± 0.017 | 0.781 ± 0.050 |
| Case 3 | 20.95 ± 3.48 | 12.60 ± 1.48 | 10.77 ± 1.17 | 0.305 ± 0.024 | 0.277 ± 0.014 | 0.281 ± 0.012 |
| Case 4 | 23.06 ± 5.04 | 16.15 ± 9.92 | 13.90 ± 2.27 | 0.374 ± 0.039 | 0.368 ± 0.023 | 0.388 ± 0.017 |

Values are mean ± sample SD across trials within each Case and fruit-count condition, after normalization by target count. Figure 11 shows the corresponding trial-level observations.

collision events from 55 to 40 relative to Case 3, a 27.3% reduction, and increased successful approaches from 49/75 to 53/75, corresponding to a gain of 5.3 percentage points. Under the evaluated configurations, Case 4 achieved fewer collision events and a higher target-approach success rate than Case 3, while requiring more execution time and TCP travel per target. Figure 11 presents the trial-level distributions of execution time, TCP path length, collision events, and approach success for each Case and fruit-count condition.

The comparison by fruit count showed the same pattern in motion efficiency: Case 3 required less time and TCP travel than Case 4, while both continuous strategies reduced these measures relative to the Home-return strategies (Table 7). Target access followed a different pattern. Approach success was lower in six-fruit scenes than in four- and five-fruit scenes, whose pooled rates were similar (Fig. 11). Moreover, Cases 1 and 2 had identical collision counts and approach-success outcomes, whereas Case 3 achieved more successful approaches at the same collision count across the separately recorded configurations. Collision counts therefore need to be considered alongside approach success when assessing physical target access.

Both continuous strategies completed all commanded trial-level sequences, although Case 4 required replanning for more target sequences than Case 3 (Table 8). Planning success remained high and mean latency was low across cases. Considered alongside the approach-success results, this shows that completion of a commanded sequence did not necessarily correspond to physical target access. The field evaluation therefore distinguishes the ability to execute the motion sequence from the ability to reach the fruit within the crop environment.

## 6. Discussions

### 6.1 Dual-arm based harvesting robot

Single-arm harvesting robots have a limited ability to cope with complex, real-world agricultural environments. They often struggle with occlusion caused by dense foliage or overlapping fruit and have restricted reachability due to their fixed base and limited degree of freedom. More critically, they display poor adaptability to dynamic disturbances, such as fruit or branch movement caused by wind, physical interaction, or the harvesting process. These disturbances can significantly degrade the performance of vision-based perception and cause failed grasping or unnecessary replanning. Thus, single-arm systems lack the robustness required for consistent performance under unstructured and fluctuating field conditions.

In contrast, dual-arm manipulation provides a more human-like solution, addressing these limitations. One arm can be dedicated to environmental interaction, such as pushing aside obstructing leaves or stabilizing moving targets, whereas the other can perform the harvesting task. This separation of roles enables the robot to maintain clear visibility of the target and increases the reliability of perception, even in the presence of external disturbances. Moreover, in scenarios where multiple pieces of fruit are spatially separated, both arms can operate simultaneously to improve efficiency. The present experiments evaluated harvest ordering and waypoint execution using the right manipulator of the dual-arm platform. Extending the framework to coordinated dual-arm operation would require task allocation and inter-arm collision avoidance, together with experimental evaluation of their effects on harvesting performance. Further, integrating tactile and visual feedback in future systems enhances the adaptability and robustness of dual-arm harvesting in dynamic agricultural environments.

### 6.2 Obstacle avoidance

Obstacle avoidance algorithms have been extensively employed in unstructured agricultural environments, enabling robots to produce collision-free paths around complex plant structures. However, although obstacle avoidance algorithms often compute feasible trajectories, their performance in dense and cluttered crop scenarios remains limited in practice. When fruit is embedded in overlapping leaves or neighboring stems, the planner may fail to determine a valid path satisfying the collision constraints and task-specific approach requirements. Even if a path is generated, it often brings the robotic end-effector or links close to nearby obstacles. This proximity can result in unintended contact with flexible plant elements (e.g., leaves or branches), causing them or the fruit to move slightly during execution.

Such environmental disturbances directly affect the perception system. Visual localization of target fruit depends on spatial consistency, and even minor movement caused by physical interaction can cause an inaccurate position estimation. In turn, this inaccurate estimation increases the likelihood of grasp misalignment or failure to complete a harvesting action. In some cases, repeated failure to generate feasible paths triggers multiple replanning attempts, increasing the motion duration and possibility of destabilizing the scene. Hence, conventional obstacle avoidance alone is insufficient in these environments. A more robust planning framework should consider geometric constraints and minimize the risk of disturbing the scene during execution. Integrating models of plant flexibility, occlusion dynamics, and real-time feedback from perception modules is essential for achieving reliable performance in field conditions.

## 7. Conclusions

This paper proposes a GA-based harvest ordering system for dense and unstructured crop environments. The proposed method combines RGB-D fruit localization, GA-based harvest ordering, and static/dynamic waypoint execution. The GA uses 3D Euclidean target distances, while the dynamic waypoint is updated from the remaining-target centroid without changing the GA sequence. The GA reached the exhaustive-search minimum in all validation runs, and both waypoint offsets were examined by sensitivity analysis. Laboratory and four- to six-fruit field experiments showed that continuous execution reduces unnecessary Home-return motion. Compared with continuous static-waypoint execution, the dynamic-waypoint strategy reduced collision occurrence and improved target-approach success while requiring more time and TCP travel, demonstrating an improvement in physical target accessibility under the evaluated field conditions. Future work will address larger environments, obstacle-aware planning, and dual-arm coordination.

### Declaration of Competing Interest

The authors declare that they have no known competing financial interests or personal relationships that could have appeared to influence the work reported in this paper.

### Acknowledgment

This work was supported by Korea Institute of Planning and Evaluation for Technology in Food, Agriculture and Forestry(IPET) through Agriculture and Food Convergence Technologies Program for Research Manpower development, funded by Ministry of Agriculture, Food and Rural Affairs(MAFRA)(RS-2024-00402136).